\documentclass[10pt, a4paper]{article}

\usepackage{lrec-coling2024} 

\usepackage{booktabs}
\usepackage{multirow}
\usepackage{pifont}
\usepackage[normalem]{ulem}
\usepackage{url}
\usepackage{enumitem}

\title{\textit{CookingSense}: A Culinary Knowledgebase \\ with Multidisciplinary Assertions}

\name{Donghee Choi$^{1 \ast}$\thanks{$^{\ast}$Most contributions to this work were made during the author's internship at Sony AI and postdoctoral tenure at Korea University.}, Mogan Gim$^{2}$, Donghyeon Park$^{3}$, Mujeen Sung$^{4}$, \\
{\bf \large Hyunjae Kim$^{2}$, Jaewoo Kang$^{2 \dagger}$, and Jihun Choi$^{5 \dagger}$\thanks{$^{\dagger}$Corresponding authors.}}}

\address{$^{1}$Imperial College London, $^{2}$Korea University, $^{3}$Sejong University, $^{4}$Kyung Hee University, $^{5}$Sony AI \\
         donghee.choi@imperial.ac.uk, \{akim, hyunjae, kangj\}@korea.ac.kr\\
         parkdh@sejong.ac.kr, 
         mujeensung@khu.ac.kr,
         jihun.a.choi@sony.com}

\abstract{
This paper introduces CookingSense, a descriptive collection of knowledge assertions in the culinary domain extracted from various sources, including web data, scientific papers, and recipes,
from which knowledge covering a broad range of aspects is acquired.
CookingSense is constructed through a series of dictionary-based filtering and language model-based semantic filtering techniques, 
which results in a rich knowledgebase of multidisciplinary food-related assertions. 
Additionally, we present FoodBench, a novel benchmark to evaluate culinary decision support systems.
From evaluations with FoodBench, we empirically prove that CookingSense improves the performance of retrieval augmented language models.
We also validate the quality and variety of assertions in CookingSense through qualitative analysis.
\\ \newline  }

\begin{document}

\maketitleabstract

\section{Introduction}
Cooking is one of the most important human activities; it not only fulfills the physiological needs of humans but also facilitates a physically and emotionally healthy life \cite{spencer2017food}.
It is intertwined with many parts of our society, including restaurant business, food manufacturing, public health, and social media \cite{lopez2015foodlab}.

Due to the importance of cooking on human beings, there have been significant amount of work that applied computational approaches into the food domain.
Especially, recent advancements in machine learning or artificial intelligence (AI) have stimulated the development of artificial intelligence AI-driven culinary decision support systems.
Since the performance of such applications is highly dependent to the existence and quality of data to be used,
it is clear that they can benefit from high-quality cooking knowledge in terms of both practicability and reliability.

\begin{figure*}
\centering
	\includegraphics[width=0.65\textwidth]{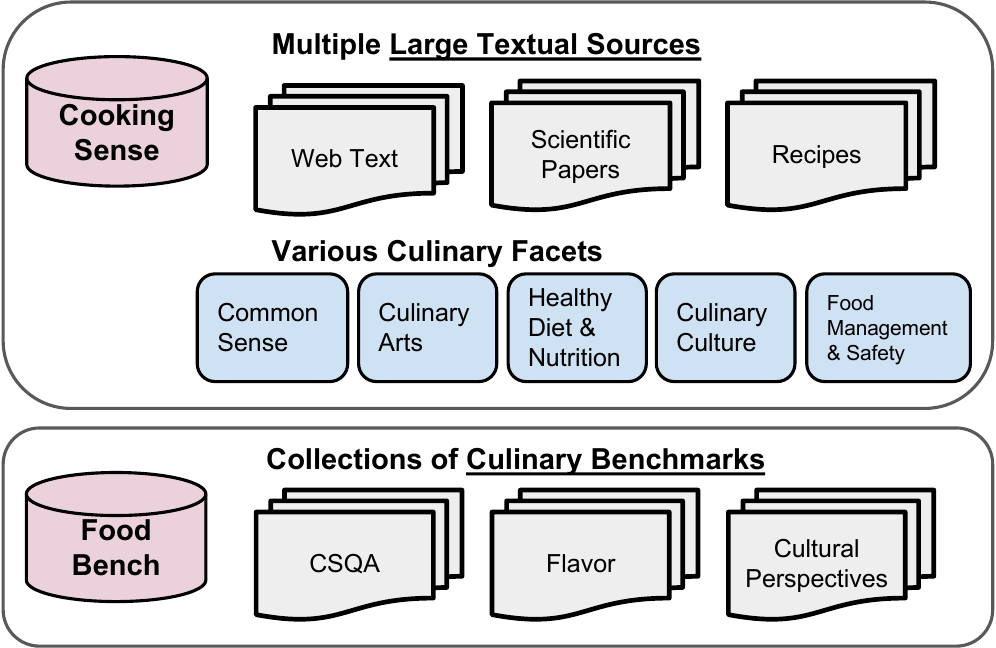}
	\caption{Our main knowledgebase, \textit{CookingSense}, is built upon diverse culinary knowledge sources. We assess the usefulness of CookingSense using the benchmark framework, \textit{FoodBench}, in assessing the capability of decision-making in the culinary domain.}
\label{fig:CookingSense-overview}
\end{figure*}

On the other hand, it is difficult to define the term `cooking knowledge' in a single sentence;
the meaning or coverage of the term could diverge depending on areas of interests and goals to pursue.
For example, for chefs who want to make savory dishes for their customers, cooking knowledge will refer to proficiency in culinary arts and in-depth understanding of food science \cite{dornenburg2008flavorbible,lopez2015foodlab}.
On the contrary, researchers working on environment, social well-being, or nutrition would focus rather on different aspects, such as environmental impact, relationship between ingredients or cooking methods and health risks \cite{marcus2013culinary_nutrition}, and so forth.

Considering this, culinary knowledge should allow individuals from various groups to derive outcomes that best suit their needs in accordance with their preferences.
In other words, cooking knowledge should be defined in a multifaceted way in order to cover a broad range of topics specialized for each group, such as food common sense \cite{wang2023sweet_commonsense}, culinary arts, health, nutrition, culinary culture \cite{nguyen2022candle}, food management, food safety, and so on.
Using resources of cooking knowledge brings several challenges when this multifaceted nature is not considered, including determining knowledge aspects to be used depending on each user's preferences, and limitation of utilizing a resource built for a specific purpose into other areas, to name a few.

Notwithstanding the importance of multifaceted cooking knowledge, many existing knowledge resources in the food domain tend to focus only on a specific aspect, e.g. recipe, nutrition, healthcare \citep{marin2019recipe1m+,fukagawa2022usda,huang2019healthcare_kg,min2022foodkg_survey}.
This justifies the motivation of constructing a large-scale, versatile cooking knowledgebase (KB) that is widely accessible and contains rich sources of food-related information.

In this work, to circumvent the lack of coverage appearing in those aspect-specific KBs, we suggest \textit{CookingSense}, a cooking KB built from various large-scale corpora.
\textit{CookingSense} is constructed through processing culinary-focused textual data from a variety of available sources with different information characteristics e.g. web data, scientific papers, and recipes, to collect descriptive knowledge statements.
We additionally create \textit{FoodBench}, a novel benchmark for assessing the capabilities of models that assist food-related decision making.
\textit{FoodBench} consists of various evaluation tasks, including flavor prediction, ingredient categorization, and culinary question answering \cite{nguyen2022candle,palta2023fork}.

We demonstrate that the performance of existing language models can be improved by incorporating \textit{CookingSense} through evaluations using \textit{FoodBench}.
Additionally, from the extensive investigation, we show \textit{CookingSense} provides richer descriptions with diverse semantics, offering wider understanding of culinary concepts.

Our contributions are summarized as follows:
\begin{itemize}
    \item We construct \textit{CookingSense}, a novel large-scale culinary KB with various cooking aspects. 
    \item We construct \textit{FoodBench}, a benchmark framework for the evaluation of culinary decision-making systems' capabilities of capturing related knowledge.
    \item We compare the effectiveness of \textit{CookingSense} against existing general and culinary domain KBs, from evaluation with recent generative language models augmented with knowledge retrieval.
\end{itemize}

We make the scripts to construct \textit{CookingSense} and \textit{FoodBench} publicly available.\footnote{\url{https://github.com/dmis-lab/cookingsense}}

\begin{table*}[]
\centering
\scalebox{1}{
\begin{tabular}{l|crrrr}
\toprule
                      & \multicolumn{1}{c}{\textbf{Relevance}} & \multicolumn{1}{c}{\textbf{Source}} & \multicolumn{1}{c}{\textbf{Coverage}} & \multicolumn{1}{c}{\textbf{Relation}} & \multicolumn{1}{c}{\textbf{Volume}} \\
                      \midrule
\textbf{ConceptNet}~ \citelanguageresource{speer2017conceptnet}   &                                      & \setlength{\tabcolsep}{1pt}\begin{tabular}{ccc} G & \hphantom{A} & \hphantom{C} \end{tabular}                                  & \setlength{\tabcolsep}{1pt}\begin{tabular}{ccccc} 1 & 2 & 3 & 4 & 5 \end{tabular}                            & Structured                                     & 980K                           \\
\textbf{FooDB}~\cite{foodb2020foodb}        & \ding{52}                                      & \setlength{\tabcolsep}{1pt}\begin{tabular}{ccc} \hphantom{G} & \hphantom{A} & C \end{tabular}                                    & \setlength{\tabcolsep}{1pt}\begin{tabular}{ccccc} \hphantom{1} & \hphantom{2} & 3 & 4 & \hphantom{5} \end{tabular}                                    & Structured                                     & 6K                                  \\
\textbf{CANDLE}~\citelanguageresource{nguyen2022candle}       &                                       & \setlength{\tabcolsep}{1pt}\begin{tabular}{ccc} G & \hphantom{A} & \hphantom{C} \end{tabular}                                   & \setlength{\tabcolsep}{1pt}\begin{tabular}{ccccc} 1 & 2 & 3 & 4 & 5 \end{tabular}                            & Textual                                     & 60K                                 \\
\textbf{Quasimodo}~ \citelanguageresource{romero2019quasimodo}    &                                      & \setlength{\tabcolsep}{1pt}\begin{tabular}{ccc} G & \hphantom{A} & \hphantom{C} \end{tabular}                                   & \setlength{\tabcolsep}{1pt}\begin{tabular}{ccccc} 1 & 2 & 3 & \hphantom{4} & \hphantom{5} \end{tabular}                               & Textual                                     & 627K                           \\
\textbf{RecipeDB}~\cite{batra2020recipedb}     & \ding{52}                                     & \setlength{\tabcolsep}{1pt}\begin{tabular}{ccc} \hphantom{G} & \hphantom{A} & C \end{tabular}                                   & \setlength{\tabcolsep}{1pt}\begin{tabular}{ccccc} \hphantom{1} & 2 & 3 & 4 & \hphantom{5} \end{tabular}                  & Structured                                    & 118K$^\ast$                               \\
\midrule
\textbf{CookingSense} (Ours) & \ding{52}                                     & \setlength{\tabcolsep}{1pt}\begin{tabular}{ccc} G & A & C \end{tabular}                               & \setlength{\tabcolsep}{1pt}\begin{tabular}{ccccc} 1 & 2& 3& 4& 5 \end{tabular}                         & Textual                                     & 54M                                \\
\bottomrule
\end{tabular}
}
\caption{
Comparison of culinary KBs.
\textbf{Relevance}: Direct relevance to culinary knowledge; 
\textbf{Source}: 
(G) General corpus, (A) Academic corpus, (C) Culinary-focused;
\textbf{Coverage}: 
Each number implies (1) Food common sense,
(2) Culinary arts,
(3) Health \& nutrition,
(4) Culinary culture,
(5) Food management \& food safety;
\textbf{Relation}: Structured indicates structured KB, while Textual is for textual KB;
\textbf{Volume}: Number of sentences in the KB 
($^\ast$: Number of recipes).
}
\label{table:compare_with_others}
\end{table*}

\section{Related Work}

Culinary KBs play a crucial role in a wide range of culinary applications, including dietary recommendation systems \cite{choi2023kitchenscale}, diet-disease management \cite{nian2021disease_discovery}, food-related question answering systems \cite{haussmann2019foodkg_recommend}, novel recipe combination recommendation \cite{park2019kitchenette, park2021flavorgraph, gim2021recipebowl,gim2022recipemind}, and more \cite{min2022foodkg_survey}.

However, current approaches to constructing culinary KBs often focus on specific data aspects, such as recipes \citelanguageresource{batra2020recipedb} or nutritional data \cite{haussmann2019foodkg_recommend}, which results in fragmented knowledge representation. 
Although individual KGs may contain substantial amounts of data, the isolation between these KGs limits the ability to gain a comprehensive overview of the overall culinary landscape.

Also, many culinary KBs are often constructed using limited lists of relations that are defined by individual researchers, resulting in lack of rich semantics and comprehensive coverage. 
Furthermore, the automatic construction of culinary KBs has mostly been done without assumption of using those with language models nor using language models (LMs) themselves in the process of KB construction, even though LMs have the potential to capture more nuanced semantics in the construction pipeline and now have being acted as the de-facto standard for natural language processing.

On the contrary, there exist a number of recent approaches to automatic construction of large KBs for other domains, such as general common sense \cite{nguyen2021ascent,nguyen2022ascentplus, hwang2021comet,bosselut2019comet}, negative relationship modeling \cite{arnaout2022uncommonsense}, and cultural perspectives \cite{nguyen2022candle}.
These approaches have made significant advancements, primarily due to the utilization of LMs \cite{arnaout2022uncommonsense,nguyen2022candle} and the enhancement of construction pipelines \citelanguageresource{bhakthavatsalam2020genericskb}.
Advancements in extractions and filtering brought by LMs enable the extraction of large volumes of data, facilitating the construction of knowledge bases in a reliable manner.
DEER \cite{huang2022deer} is one of noteworthy approaches in this line, which proposed KBs with descriptive relationships that contain rich semantics among a set of concepts.
Our work aligns with the recent advancements in previous studies, presenting effective pipelines for building reliable and large-scale KBs in the culinary domain.

\section{KB Construction}
\label{sec:kb-construction}

\begin{figure*}
\centering
	\includegraphics[width=0.9\textwidth]{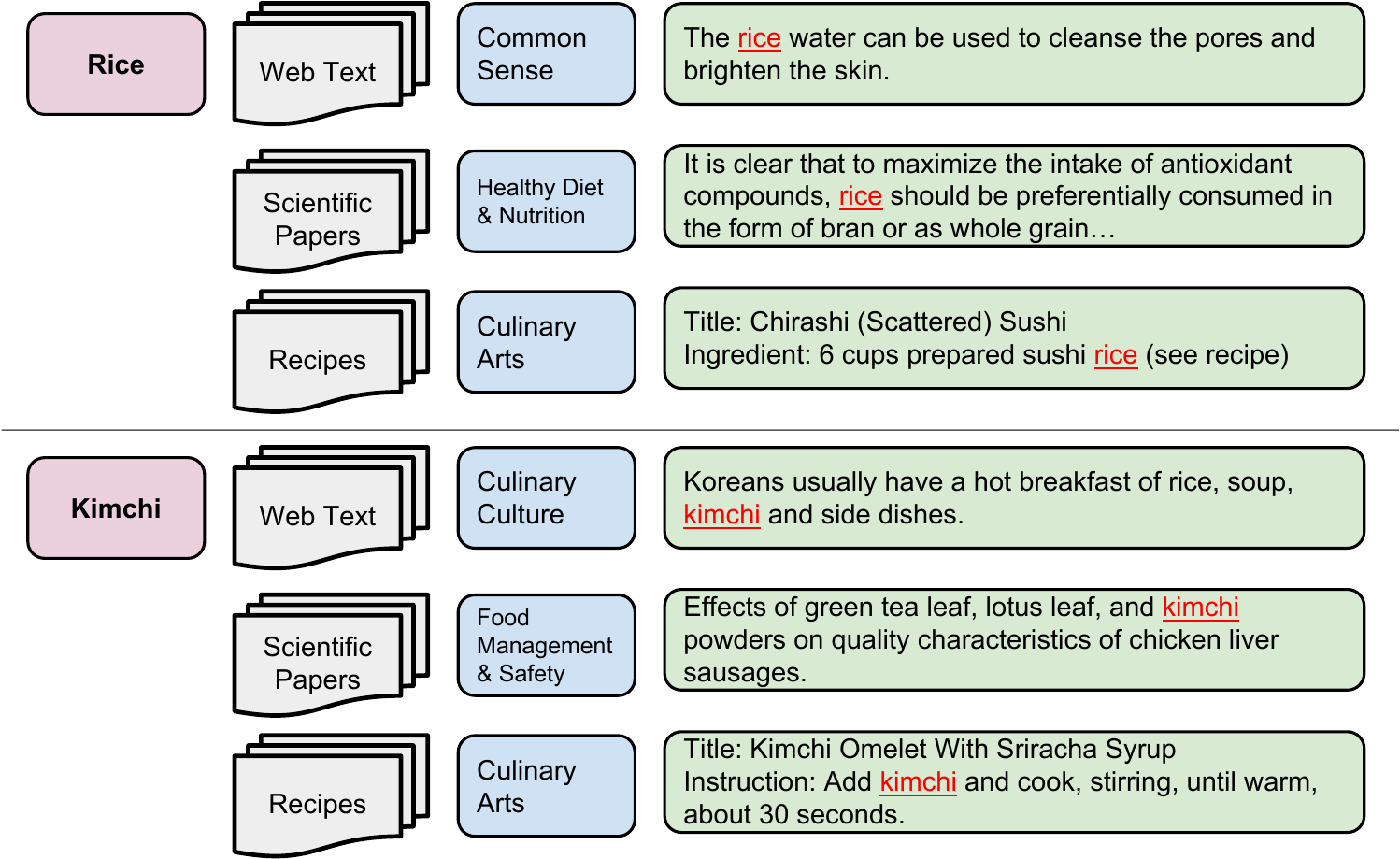}
	\caption{
 \textit{CookingSense} examples.  
 We present a selection of examples representing two culinary concepts, such as \textcolor{red}{\underline{rice}} and \textcolor{red}{\underline{kimchi}}, across various types in Table~\ref{table:CookingSense_categorization_results}.}
\label{fig:CookingSense-example}
\end{figure*}

We now provide a detailed description about the processes for constructing the \textit{CookingSense} dataset.
To ensure broad coverage and collect high-quality knowledge statements within the domain, we chose three types of text corpora: \textit{web data}, \textit{scientific papers}, and \textit{recipes}.

\begin{figure*}
\centering
	\includegraphics[width=0.62\textwidth]{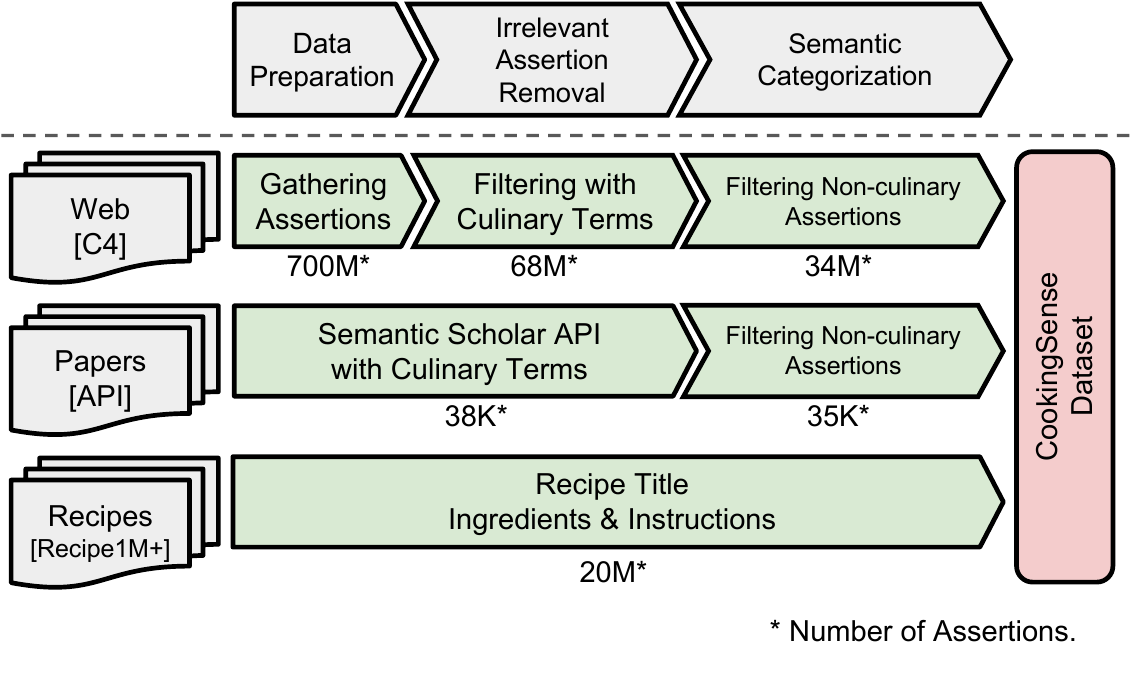}
	\caption{Pipeline of CookingSense KB construction.}
\label{fig:CookingSense-pipeline}
\end{figure*}

While these corpora accompany valuable information across various aspects, there may also exist texts with undesired or no information, e.g. noisy texts, texts not in the culinary domain.
To filter out those texts, we apply various filters, each of which is designed based on linguistic properties, thesaurus of words, semantics of knowledge statements, and so forth.
Figure~\ref{fig:CookingSense-example} depicts the example knowledge statements from \textit{CookingSense}, and Figure~\ref{fig:CookingSense-pipeline} illustrates the overview of the pipeline for the KB construction along with the number of assertions before and after each step.

\subsection{Requirements}

We elaborated the following requirements to ensure the reliability and applicability of CookingSense in various cooking-related downstream tasks.

\begin{itemize}
    \item \textbf{Relevance to cooking:}
    Relevance to cooking is one of the most important criteria we want to achieve.
    Resources used, construction pipeline, and benchmark tasks should have relevance to the culinary domain.

    \item \textbf{Multiple data sources:}
    The KB should obtain knowledge from various sources,
    to incorporate knowledge that is not biased towards a specific community or content.
    
    \item \textbf{Coverage on various facets:}
    In line with the above requirement, the KB should contain knowledge that can cover various use cases and preferences, i.e. multifaceted nature.
    Unlike many existing culinary datasets that predominantly focus on recipes or general corpus, CookingSense encompasses various aspects of culinary knowledge by acquiring knowledge from web content, paper-based corpora, and user-generated recipes.
    
    \item \textbf{Flexible relation types:}
    An optimal set for relation types required by a task set could differ, depending on the goal for each application.
    CookingSense prioritizes inclusion of a wide range of culinary semantics represented in a form of text,
    by employing unsupervised mining techniques to capture semantic variety.
    It employs unsupervised mining techniques to capture this semantic variety.
    
    \item \textbf{Sufficiently large volume:}
    Even though there exist numerous factors that determine the usability of a KB, size is one of the most important components relevant to coverage, diversity, correctness, and robustness.
    CookingSense is designed to have a substantial amount of knowledge assertions.
\end{itemize}

We compare CookingSense against other baseline datasets upon these requirements in Table~\ref{table:compare_with_others}.

\subsection{Data Sources}
\label{subsec:data-sources}

We chose three different data sources: web data, scientific papers, and recipes, as base corpora to construct CookingSense.
Data sources are primarily composed of English texts.

\subsubsection{Web Data}
    We used Colossal Clean Crawled Corpus \citep[C4;][]{raffel2020t5} for our base corpus for web data.
    C4 is a large-scale web corpus consisting of about 364M articles (7B sentences) from the web.
    Due to its massive size and wide range of coverage, we chose C4 for our key data source for general culinary and food-related information.
    Although several noise reduction procedures, including removing harmful texts,  were done in the process of constructing C4, there still exist a significant amount of texts which are noisy or not in our interest, we applied a series of data refinement techniques on the original C4.

\subsubsection{Scientific Papers}
    We chose scientific papers as another source for our KB, in the belief that it enables us to integrate trustworthy and research-backed knowledge into our KB.
    We collected a large amount of scientific literature using Semantic Scholar Public API\footnote{\url{https://www.semanticscholar.org/product/api}}.
    These APIs grant access to a vast collection of academic articles;
    Semantic Scholar Public API provides access to S2ORC \citep{lo2020s2orc}, a large corpus of 81.1M open-access academic papers from various fields.
    
    To retrieve papers relevant to our interest, we built a list of terms in the domains of culinary arts, nutrition, and food sciences and used them to query the APIs.
    Detailed descriptions of those terms are available in \S\ref{subsec:term-filter}.
    For each retrieved article, we collected the title and the summary of the abstract. 

\subsubsection{Recipes}
    Recipes could also be a useful source for a food-related KB, in that a recipe usually contains procedural knowledge required for making a dish from basic ingredients.
    We used Recipe1M+ \cite{marin2019recipe1m+} for our data source for recipes.
    Recipe1M+ consists of more than 1M culinary recipes with their pictures gathered from a large number of popular recipe websites.
    We extracted the title, the list of ingredients, and cooking instructions from each recipe.



\subsection{Creation of Assertions}
Depending on the inherent characteristics of each data source, the length of each text instance (i.e. document, paragraph, or sentence) may vary.
To address this discrepancy, we split or merge text instances into chunks of one or two sentences and use them as the unit of knowledge; which we refer to as ``assertions.''

\begin{itemize}
    \item For C4, we utilize the sentence tokenizer of spaCy \citep{spacy2} and treat each sentence as an assertion.
    \item For the scientific papers, there exist two types of texts: 
    Since the output of SciTLDR tends to be a single sentence, we concatenate the title and the summary of the paper generated by SciTLDR to compose an assertion.
    \item In Recipe1M+, the content of each recipe consists of the section of ingredient explanations and the section of cooking instructions.
    We combine the recipe title and either a single ingredient or a single step of cooking instruction to put together an assertion.
\end{itemize}

The average number of tokens in each assertion by data source is available at Table~\ref{table:token_len}.

\begin{table}[htb]
\centering
\scalebox{0.9}{
\begin{tabular}{l|r}
\toprule
\textbf{Source} & \textbf{Avg. Length} \\
\midrule
\textbf{Web} & 16.96 \\
\textbf{Paper} & 48.71 \\
\textbf{Recipe} & 11.50 \\
\bottomrule
\end{tabular}
}
\caption{Average token length of each assertion by its data source.}
\label{table:token_len}
\end{table}

\subsection{Removal of Non-Generic Assertions}
Not all the assertions collected from the above process contain knowledge that can be accepted as true in general.
We apply filtering on assertions to remove non-generic assertions, which include non-informative or context-dependent statements.
Following approaches used in constructing GenericsKB \citep{bhakthavatsalam2020genericskb} and CANDLE \citep{nguyen2022candle} with identical purpose to ours, we make use of 27 handcrafted rules used in GenericsKB \citep{bhakthavatsalam2020genericskb} to automatically filter out non-generic assertions. 

The rules are defined under various assumptions on generic sentences, including the ones based on parse trees (e.g. removing sentences whose root is non-verb), modals (e.g. removing sentences containing `could', `would'), first word (e.g. remove sentences starting with a determiner `a', `the').

\begin{table}[]
\centering
\scalebox{0.9}{
\begin{tabular}{l|l|l}
\toprule
\multicolumn{1}{c}{\textbf{Food Name}} & \multicolumn{1}{c}{\textbf{Ingredient Name}} & \multicolumn{1}{c}{\textbf{Other Terms}} \\
\midrule
crock pot                              & salt                                         & vitamin                                  \\
black bean                             & onion                                        & mineral                                  \\
italian sausage                        & butter                                       & protein                                  \\
french toast                           & water                                        & fat                                      \\
roast beef                             & garlic clove                                 & carbohydrate                             \\
\bottomrule
\end{tabular}
}
\caption{Example words for the irrelevant cooking knowledge
assertion filter described in \S\ref{subsec:term-filter}.}
\label{table:CookingSense_wordfilter}
\end{table}

\begin{table*}[]
\centering
\scalebox{1.0}{
\begin{tabular}{l|r|r|r|r}
\toprule
 \multicolumn{1}{c}{\textbf{Type Description}}
                & \multicolumn{1}{c}{\textbf{Web}}                         & \multicolumn{1}{c}{\textbf{Paper}}
                & \multicolumn{1}{c}{\textbf{Recipe}}
                 & \multicolumn{1}{c}{\textbf{Total}}\\
                \midrule
 Food Common Sense & 7,210,883 & 3,262  & - &7,214,145                          \\
Culinary Arts & 5,630,583                                                & 703       & 20,372,992   &26,004,278                  \\
Healthy Diet \& Nutrition & 6,211,601                                                & 21,317              & -  &6,232,918             \\
Culinary Culture & 4,414,988                                                & 212    & -  &4,415,200                           \\
Food Management \& Food Safety & 10,846,348                                               & 9,596  & -        &10,855,944                   \\
\midrule
All & 34,314,403	&35,090	&20,372,992	& 54,722,485 \\
\bottomrule

\end{tabular}}
\caption{
Number of assertions in \textit{CookingSense}: distribution by types and sources.
}
\label{table:CookingSense_categorization_results}
\end{table*}



\subsection{Removal of Irrelevant Assertions}
\label{subsec:term-filter}

After the removal of non-generic assertions, we now have a collection of generic assertions.
However, the remaining assertions still have a wide variety of content and perspectives that reside beyond our specific area of focus.
To address this, we designed a filtering method based on the dictionary of food or culinary terms we collected.
This filter eliminates assertions that are generic yet irrelevant to our target domain.

Our filtering dictionary primarily consists of two types of terms: (1) ingredient and food name and (2) food-related terms obtained from an AI assistant.
Specific examples are provided in Table~\ref{table:CookingSense_wordfilter} for reference.

\subsubsection{Ingredient and Food Names}
    We use RecipeDB \cite{batra2020recipedb} in collecting entities associated with food names and ingredient names. 
    We extract bigrams appearing in recipe titles to obtain the list of food names, and bigrams appearing in ingredient sections to obtain the list of ingredient names.
    Bigrams that occur more than 3 times (food names) or 2 times (ingredient names) to avoid rare or noisy entities to be included in the dictionary; 1,914 and 5,482 bigrams are collected as a result.
    We show the most frequent bigrams for each data source in Table~\ref{table:bigram_stats_all}.

\subsubsection{Terms Collected from AI Assistant}
    Relying only on ingredient and food names extracted from a recipe database would limit the ability to capture a broader range of culinary terms, since there could exist food-related assertions that do not necessarily contain those terms, which in result may narrow down the scope of the resulting assertions.
    
    To mitigate this, we also add general terms such as ``Food'' and ``Nutrition'' as well as specific terms such as ``Vitamin B'' for nutrition and ``Diabetes'' for healthy diet.
    We develop a two-level ontology, allowing us to categorize 1,600 culinary terms.

    We use ChatGPT\footnote{\url{https://chat.openai.com}} to collect those common culinary terms.
    The prompt ``Please provide an exhaustive list of verbs related to cooking actions and techniques, such as chopping, slicing, seasoning, and garnishing.'' is used to collect common culinary terms.
    Additionally, to acquire the list of professional or scientific terms related to food, we utilize the prompt ``I want to develop a dataset based on food computing and want to aggregate abstracts of related papers. Please suggest me 20 keywords that will provide such insights.''\footnote{This prompt is also used in collecting terms for making S2ORC queries, as described in \S\ref{subsec:data-sources}.}

\begin{table}[]
\scalebox{0.9}{
\begin{tabular}{l|l|l}
\toprule
\multicolumn{1}{c}{\textbf{Web}} & \multicolumn{1}{c}{\textbf{Paper}} & \multicolumn{1}{c}{\textbf{Recipe}} \\
\midrule
ice cream                       & fatty acids                        & olive oil                           \\
olive oil                       & gut microbiota                     & teaspoon salt                       \\
hot water                       & systematic review                  & black pepper                        \\
weight loss                     & oxidative stress                   & brown sugar                         \\
blood sugar                     & antioxidant activity               & finely chopped                      \\
\bottomrule
\end{tabular}
}
\caption{Top 5 bigrams by frequencies in \textit{CookingSense} by its sources.}
\label{table:bigram_stats_all}
\end{table}
\begin{table*}[]
\centering
\scalebox{0.72}{
\begin{tabular}{c|l|l|l|l|l}
\toprule
\multicolumn{1}{l}{\textbf{}}    & \multicolumn{1}{c}{\textbf{Food Common Sense}} & \multicolumn{1}{c}{\textbf{Culinary Arts}} & \multicolumn{1}{c}{\textbf{Health \& Nutrition}} & \multicolumn{1}{c}{\textbf{Culinary Culture}} & \multicolumn{1}{c}{\textbf{Food Management \& Safety}} \\

\midrule 
\multirow{5}{*}{\textbf{Web}}    & ice cream                           & olive oil                           & weight loss                         & new year                            & hot water                           \\
                                 & dining room                         & ice cream                           & blood sugar                         & ice cream                           & drinking water                      \\
                                 & living room                         & stainless steel                     & vitamin c                           & new york                            & water quality                       \\
                                 & dining area                         & white wine                          & blood pressure                      & united states                       & water damage                        \\
                                 & peanut butter                       & lemon juice                         & vitamin d                           & world famous                        & water supply                        \\

\midrule
\multirow{5}{*}{\textbf{Paper}} & food security                       & soy sauce                           & fatty acids                         & food pairing                        & food safety                         \\
                                 & climate change                      & fish sauce                          & gut microbiota                      & medicinal plants                    & food waste                          \\
                                 & genetic diversity                   & alcoholic fermentation              & systematic review                   & cultural food                       & public health                       \\
                                 & genome sequence                     & lactic acid                         & oxidative stress                    & flavor network                      & escherichia coli                    \\
                                 & fruit ripening                      & acid bacteria                       & mediterranean diet                  & flavor compounds                    & listeria monocytogenes              \\
\bottomrule
\end{tabular}
}
\caption{
Top 5 bigrams from CookingSense by types.
}
\label{table:bigram_stats_by_types}
\end{table*}

\subsection{Semantic Categorization}
\label{subsec:semantic-filter}

To make the KB more usable and figure out which category each assertion falls into, we constructed a ``silver standard'' annotated dataset where category labels related to culinary arts and food-related content are attached to assertions using a large language model (LLM).

For the initial dataset, we randomly sampled 10,000 assertions from the KB gathered through the method described previously.
To annotate labels on those 10,000 assertions, we use the GPT-4 model\footnote{\url{https://openai.com/research/gpt-4}} (version as of September 30, 2023) to classify them into six distinct types: (a) Food Common Sense, (b) Culinary Arts, (c) Healthy Diet \& Nutrition, (d) Culinary Culture, (e) Food Management \& Food Safety, and (f) Irrelevant or None. 

The distribution across these categories exhibited significant imbalance, where the majority of sentences fell into the ``Irrelevant or None'' category. 
To address this class imbalance issue and facilitate classifier training, we employed a well-established data-level approach---under-sampling the dataset \cite{johnson2019survey_classimbalance}.
Specifically, we randomly chose 218 sentences from each category, resulting in a balanced dataset. 
This balanced dataset is subsequently divided in the ratio of 80\% and 20\%, each for training and test set. 

After that, we trained a classification model based on the \texttt{bert-large-uncased} architecture \cite{devlin-etal-2019-bert} using the training split of the balanced dataset.
It achieved an accuracy of 0.76 on the test split, underscoring the model's efficacy in categorizing assertions. 
We applied this classifier to 68M assertions after removing irrelevant assertions, resulting in 34M categorized assertions, excluding those labeled as ``Irrelevant or None.'' Detailed results are available in Table \ref{table:CookingSense_categorization_results}.

In addition, we present bigrams with the highest frequencies categorized by their sources and types in Table~\ref{table:bigram_stats_all} and \ref{table:bigram_stats_by_types}. 
The distributions of bigrams demonstrate that information varies across different sources and types, highlighting the importance of collecting data from diverse sources.

\section{Evaluation}

To assess the effectiveness of our KB, we adopt the context-augmented language model setup inspired by the work of Retrieval Augmented Generation \cite[RAG;][]{lewis2020rag},
where a context retrieved from a retriever system is augmented with the input to generate texts.
We use baseline KBs and the CookingSense as sources for retrieval to measure how differently knowledge assertions from other KBs enrich the input.

\textbf{Retriever system}: 
We adopt Okapi-BM25 \cite{robertson2009bm25} for the retriever system for RAG evaluation, using the retriv.\footnote{\url{https://github.com/AmenRa/retriv}}
BM25 is a simple yet powerful ranking algorithm based on term and document frequency, which is widely used in various work \cite{trotman2014improvements}.
The motivation behind choosing BM25 for our retriever is, to keep a retrieval algorithm as simple as possible so that the generation quality depends more on KB's quality, not the performance of a retrieval algorithm.

\textbf{Language model}:
We utilized the Flan-T5 (\texttt{flan-t5-large}) language model \cite{chung2022flant5} for text generation purposes. 
Flan-T5 is a language model based on T5 \cite{raffel2020t5} fine-tuned with instruction guides \cite[\textit{inter alia}]{wei2022finetuned,ouyang2022training,sanh2022multitask}. 
It can respond to a wide range of question types without the need for additional fine-tuning specific to a benchmark format. 

\subsection{FoodBench}

To evaluate the utility of CookingSense and other baseline KBs, we have developed a benchmark suite for the culinary domain, namely \textit{FoodBench}.
FoodBench is a collection of culinary-related benchmark tasks covering question answering, flavor perspective prediction, and cultural perspective prediction.
To ensure compatibility within our RAG framework, we converted these tasks into a multiple choice question answering format.
For instance, in a question like ``What type of cut does something that is minced produce?" with answer choices ``a) squares, b) long strips, c) large slices, d) very tiny pieces", the agent's task is to select the correct answer, which, in this case, would be d).
Also in our evaluation, if the number of potential answers for a question is less than four, we designate any remaining possibilities as ``This is not an answer."

\begin{table*}[]
\centering
\scalebox{1}{
\begin{tabular}{l|ccccc|r}
\toprule
                         & \multicolumn{1}{c}{\textbf{CSQA}} & \multicolumn{1}{l}{\textbf{ASCENT++}} & \multicolumn{1}{l}{\textbf{TGSC}} & \multicolumn{1}{l}{\textbf{CKQ}} & \multicolumn{1}{l}{\textbf{FORK}} &
                         \multicolumn{1}{l}{\textbf{Avg.}}
                         \\
\midrule                        
\textbf{Without Context} & 16.08                                      & 24.52                                    & 13.60                                     & 14.38                               & 28.80                  & 19.48           \\
\midrule
\textbf{ConceptNet}~\citelanguageresource{speer2017conceptnet}      & 47.79                                      & 22.90                                    & 47.60                                     & 54.25                               & 46.20           & 43.75                  \\
\textbf{FooDB}~\citelanguageresource{foodb2020foodb}           & 48.25                                      & 20.97                                    & 45.80                                     & 52.29                               & \textbf{58.70}               & 45.20     \\
\textbf{CANDLE}~\citelanguageresource{nguyen2022candle}          & 48.48                                      & 41.29                                    & 51.40                                     & 54.58                               & 39.67                 & 47.08            \\
\textbf{Quasimodo}~\citelanguageresource{romero2019quasimodo}       & 50.35                                      & 40.65                                    & 63.40                                     & 53.59                               & 53.80                 & 52.36            \\
\midrule
\multicolumn{5}{l}{\textit{\ul{CookingSense}}} \\[0.10cm]
\textbf{Paper}       & 51.52                                      & 20.32                                    & 52.00                                     & 52.29                               & \ul{ 57.07}                       \\
\textbf{Recipe}      & 56.88                                      & 54.84                                    & \textbf{68.60}                            & 51.63                               & 50.00                             \\
\textbf{Web}         & \textbf{70.63}                             & \textbf{59.35}                           & \ul {65.80}                               & \textbf{66.99}                      & 51.09                             \\
\textbf{All}         & \ul {68.30}                                & \ul {56.77}                              & 65.40                                     & \ul{ 64.38}                         & 50.00                            \\
\textbf{CookingSense} (Ours)         & \textbf {68.30}                                & \textbf {56.77}                              & \textbf{65.40}                                     & \textbf{ 64.38}                         & 50.00    & \textbf{60.97}                        \\
\bottomrule
\end{tabular}
}
\caption{
Experimental results for \textit{FoodBench}.
The \textbf{bold} values indicate the highest scores within each benchmark dataset.
All scores represent result accuracy.
}
\label{table:foodbench_results}
\end{table*}

\textbf{Question answering}:
We collected 429 question-answer pairs from user-generated content on the web that reflects the real-world perspective of culinary knowledge; namely \textit{CookingSenseQA} (CSQA).

\textbf{Flavor perspectives}: 
Flavor is one of the most important feature that determines the overall experience of a dish.
We gathered and constructed flavor-related binary classification problems from the following resources:

\begin{itemize}
    \item {\textbf{ASCENT++} \citelanguageresource{nguyen2022ascentplus}}:
    ASCENT++ is a common sense KB with a diverse range of facets, including culinary concepts, and their corresponding assertions.
    We gathered 310 assertions where ingredients are associated with flavor expressions from ASCENT++.
    These assertions include an example such as \texttt{(Carambola, sweet)}.

    \item {\textbf{The Good Scents Company}}:
    Another data source we used for gathering flavor information is The Good Scents Company Information System\footnote{\url{http://www.thegoodscentscompany.com/}} (TGSC). 
    From this resource, we chose 500 assertions in a broader spectrum of flavor expressions, such as \texttt{(orange, citrus)} and \texttt{(irish cream, melon)}.
    
\end{itemize}

\begin{table*}[]
\centering
\scalebox{0.73}{
\begin{tabular}{l|l|l}
\toprule
\multicolumn{1}{c}{\textbf{Source}} & \multicolumn{1}{c}{\textbf{Question}}            & \multicolumn{1}{c}{\textbf{Retrieved Context}}                                               \\
\midrule
\multirow{2}{*}{Web}                & If you double your recipe,                       & \multirow{2}{*}{Recipe can be doubled but don’t double the salt in the cooking water.}       \\
                                    & what ingredient should you not double?           &                                                                                              \\
                                    \midrule
\multirow{2}{*}{Recipe}             & “Soft Ball” Stage of Cooked Sugar                & \multirow{2}{*}{Barley Sugar Cook to 240F or soft-ball stage.}                               \\

                                    & occurs in which temperature range?               &                                                                                              \\
                                    \midrule
Paper                               & What can you use as a substitute for real sugar? & ...alternatives to sugar with special consideration of xylitol.                              \\
\midrule
\multirow{2}{*}{Web}                & The forest in France whose oak trees are used to & \multirow{2}{*}{The wine is then distilled and given to age in French Limousin oak barrels.} \\
                                    & make barrels for aging wine is known as the:     &                                                                                              \\
                                    \midrule
\multirow{2}{*}{Web}                & Which of the following ingredients is            & Milk is considered one of the eight major food allergens by the FDA.                         \\
                                    & not considered a major eight allergen?           & Caution: nuts and peanuts are two of the top eight major food allergens.                 \\

                                \bottomrule
\end{tabular}
}
\caption{Examples of CSQA in \textit{FoodBench} with retrieved contexts from \textit{CookingSense}.
}
\label{table:qa_anaylsis}
\end{table*}

\textbf{Cultural perspectives}:
Cultural perspectives also play a crucial role in shaping culinary decision-making processes, as they influence not only the ingredients and techniques but also the traditions and rituals associated with food. 
We integrate two distinct benchmark datasets into FoodBench to cover those cultural dimensions:

\begin{itemize}
    \item {\textbf{Cultural knowledge quizzes}}: 
    We use the collection of cultural knowledge quizzes which used in the evaluation of CANDLE \citelanguageresource{nguyen2022candle}. 
    Throughout this paper, we denote this evaluation dataset as CKQ.
    It contains 500 multiple-choice questions related to cultural knowledge, and among which we chose 306 food-related question-answer pairs,
    for example ``In many European countries, which meat is consumed on Easter Sunday?''
    These question-answer pairs could be used to measure whether a KB covers diverse cultural practices and culinary traditions around the world.
    
    \item {\textbf{FORK}~ \citelanguageresource{palta2023fork}}: 
    FORK is a manually-curated dataset comprising 184 question-answer pairs designed to probe cultural biases and assumptions in the culinary domain. 
    This dataset requires cultural nuances in culinary practices through questions such as, ``A man went to a restaurant and ordered Sweet and Sour Pork. As he put some of the food in his bowl to eat, he reached out for what?"
\end{itemize}

\subsection{Baselines}
We compare CookingSense with the following KBs based on how each KB contributes to performance improvement in the FoodBench evaluation.

\begin{itemize}
    \item \textbf{ConceptNet}~ \citelanguageresource{speer2017conceptnet}: 
    ConceptNet is a structured semantic network that has been steadily improved through crowdsourcing since 1999. 
    To make it usable within our evaluation framework, we convert triples in ConceptNet (subject entity, relation type, object entity) into 980k assertions.

    \item \textbf{FooDB}~\citelanguageresource{foodb2020foodb}:
    FooDB is a structured KB that focuses on food constituents, chemistry, and biology.
    We extracted 6,059 culinary-domain knowledge snippets from this KB and converted them into assertions.
    
    \item \textbf{CANDLE}~ \citelanguageresource{nguyen2022candle}:
    CANDLE is a cultural common sense KB, spanning various facets such as food, behaviors, rituals, and traditions.
    We obtained 60,134 assertions in the culinary domain from this KB.
    
    \item \textbf{Quasimodo}~ \citelanguageresource{romero2019quasimodo}:
    Quasimodo is an open-source common sense KB designed to retrieve properties relevant to entities, including those in culinary topics.
    We gathered about 6.3M assertions extracted from the triplets within the KB.  
    This corpus functions as a large-scale, general textual knowledge resource for our evaluation.
    
\end{itemize}

\subsection{Experimental Results}
\subsubsection{FoodBench}
Table \ref{table:foodbench_results} presents the results of RAG experiments with FoodBench.
For scores from CookingSense, along with the overall performance, we also denote scores where only a specific source is used, to see each data source's effectiveness separately.

In all experiments, even with the use of recent LLM which is believed to have world knowledge with the power of massive pre-training, it is shown that integrating KB improves performance significantly.
This validates utilizing external KBs is still one of the most effective and realizable ways to improve performance, 
strengthening the necessity of a KB that contains high-quality assertions and is easy to use along with LLMs.

In most cases, RAG integrated with CookingSense outperformed other baseline KBs in various evaluation datasets by a large margin.
For the FORK, RAG with FooDB performed the best,
which we conjecture due to the fact that FooDB contains background knowledge of ingredients directly aligning with problems in FORK.

Experimental results demonstrate the potential usefulness of CookingSense in various culinary-related downstream tasks.
We expect CookingSense to provide a foundational basis for empowering other types of large language models with specific culinary knowledge to facilitate better practicability when deployed in culinary decision support systems.

\subsubsection{Qualitative Analysis}
In the previous evaluation, we used FoodBench, an automatically constructed benchmark data from available sources to show the effectiveness of CookingSense.
To verify the quality of CookingSense in a more direct and fine-grained way, we conducted a qualitative analysis of the results from CSQA experiments.
Table \ref{table:qa_anaylsis} presents a selection of question-answer pairs from CSQA along with their retrieved context form its sources.

\textbf{Finding 1: \textmd{\textit{Web data contains diverse and long-tailed information.}}}
Upon analysis, we found that due to diversity of data from the web, retrieval of common sense (row 1; it is known that it should be adjusted to taste), cultural perspectives (row 4, `French Limousin oak'), and authoritative information (row 5, statement from FDA).

\textbf{Finding 2: \textmd{\textit{Recipes offer culinary insights, while papers do expert-level knowledge.}}}
Assertions from recipes show its strength of covering empirical knowledge, including examples like `soft ball stage of cooked sugar occurs at the temperature of 240 \textdegree F' (row 2).
On the other hand, assertions from scientific papers give scientific knowledge such as `xylitol can be used as a substitute for sugar' (row 3).

In summary, these examples show that CookingSense contains a wide array of information, originating from various sources that provide rich textual representations of assertions in a different aspect, resulting in a complementarily gathered collection of knowledge.

\section{Conclusion}
In conclusion, we have constructed the CookingSense, a large-scale KB that encompasses a comprehensive collection of culinary-domain assertions obtained from various data sources.
Leveraging dictionary-based filtering and language model-based semantic filtering techniques, we obtained a collection of high-quality assertions with broad coverage in the culinary domain.
We also introduced the FoodBench benchmark framework for assessing culinary-domain decision supporting systems.

From evaluations with FoodBench, we empirically proved that CookingSense improves the performance of retrieval augmented language models.
We conducted a qualitative analysis to validate the quality and variety of assertions in CookingSense.
We expect CookingSense and FoodBench to pave the way for future work on building, enhancing, and evaluating culinary decision supporting systems.

For future work, we aim to enhance the system into a new culinary domain-specific QA system or chatbot~\cite{zhang2023domainchatbot} covering diverse perspectives related to culinary decision making using LLMs~\cite{touvron2023llama2,jiang2023mistral, team2024gemma} and the prompt engineering techniques~\cite{wei2022cot, nori2023medprompt}. 
Also, we plan to enhance our datasets with enhanced extraction techniques~\cite{hayati2023diversity,cegin2023chatgpt_diversity} that utilize recent LLMs to cover more diverse topics.

\section{Ethical Considerations and Limitations}

\subsection{Ethical Considerations}

Given that our KB and benchmark framework are created using an automated pipeline, we acknowledge a potential risk of inclusion of biased or violent data from various sources, such as web-crawled content, papers, and recipes. This introduces certain ethical considerations and limitations that need to be addressed.

Especially, biases could exist in the constructed KB and also for the benchmarks, mingled with other perspectives such as culture, ethnicity, or gender.

\subsection{Limitations}
Although we aimed to include as diverse data source as possible, the KB still has room for improvement by incorporating more diverse data sources,
as seen in the FORK experiment where FooDB helped the most while for all other experiments CookingSense improved the performance by a large margin.
We have plans to extend CookingSense by incorporating an even broader range of facets related to food and culinary arts,
such as USDA Food and Nutrition,\footnote{\url{https://www.usda.gov/topics/food-and-nutrition}} USDA FoodData Central,\footnote{\url{https://fdc.nal.usda.gov}} and FooDB.

Also, the data sources we used are in English, which may hinder collection of resources from low-resource languages, reflection of cultural nuance, and so forth.

\section{Acknowledgement}

We thank Hoonick Lee and David Im for their invaluable assistance.
Our work is part of a collaboration between Sony AI and Korea University.
This work was supported by the National Research Foundation of Korea (NRF-2023R1A2C3004176, NRF-2022R1F1A1069639, NRF-2022R1C1C1008074) and the ICT Creative Consilience Program through the Institute of Information \& Communications Technology Planning \& Evaluation (IITP) grant funded by the Korean government (MSIT) (IITP-2024-2020-0-01819, No.RS-2022-00155911 (Artificial Intelligence Convergence Innovation Human Resources Development (Kyung Hee University)).

Donghee Choi is additionally supported by the Horizon Europe project CoDiet. The CoDiet project is funded by the European Union under Horizon Europe grant number 101084642. CoDiet research activities taking place at Imperial College London and the University of Nottingham are supported by UK Research and Innovation (UKRI) under the UK government's Horizon Europe funding guarantee (grant number 101084642).

\nocite{*}
\section{Bibliographical References}\label{sec:reference}

\bibliographystyle{lrec-coling2024-natbib}
\bibliography{lrec-coling2024}

\begin{thebibliography}{8}
\expandafter\ifx\csname natexlab\endcsname\relax\def\natexlab#1{#1}\fi

\bibitem[{Batra et~al.(2020)Batra, Diwan, Upadhyay, Kalra, Sharma, Sharma, Khanna, Marwah, Kalathil, Singh, Tuwani, and Bagler}]{batra2020recipedb}
Batra, Devansh and Diwan, Nirav and Upadhyay, Utkarsh and Kalra, Jushaan Singh and Sharma, Tript and Sharma, Aman Kumar and Khanna, Dheeraj and Marwah, Jaspreet Singh and Kalathil, Srilakshmi and Singh, Navjot and Tuwani, Rudraksh and Bagler, Ganesh. 2020.
\newblock \href {https://doi.org/10.1093/database/baaa077} {\emph{{RecipeDB}: A resource for exploring recipes}}.
\newblock Oxford Academic.
\newblock PID \href{https://cosylab.iiitd.edu.in/recipedb/}{https://cosylab.iiitd.edu.in/recipedb/}.

\bibitem[{Bhakthavatsalam et~al.(2020)Bhakthavatsalam, Anastasiades, and Clark}]{bhakthavatsalam2020genericskb}
Bhakthavatsalam, Sumithra and Anastasiades, Chloe and Clark, Peter. 2020.
\newblock \href {https://arxiv.org/abs/2005.00660} {\emph{{GenericsKB}: A knowledge base of generic statements}}.
\newblock PID \href{https://allenai.org/data/genericskb}{https://allenai.org/data/genericskb}.

\bibitem[{{FooDB}(2020)}]{foodb2020foodb}
{FooDB}. 2020.
\newblock \emph{{FooDB} Version 1.0}.
\newblock PID \href{https://foodb.ca/}{https://foodb.ca/}.

\bibitem[{Nguyen et~al.(2022{\natexlab{a}})Nguyen, Razniewski, Romero, and Weikum}]{nguyen2022ascentplus}
Nguyen, Tuan-Phong and Razniewski, Simon and Romero, Julien and Weikum, Gerhard. 2022{\natexlab{a}}.
\newblock \href {https://doi.org/10.1109/TKDE.2022.3206505} {\emph{Refined commonsense knowledge from large-scale web contents}}.
\newblock PID \href{https://ascentpp.mpi-inf.mpg.de/}{https://ascentpp.mpi-inf.mpg.de/}.

\bibitem[{Nguyen et~al.(2022{\natexlab{b}})Nguyen, Razniewski, Varde, and Weikum}]{nguyen2022candle}
Nguyen, Tuan-Phong and Razniewski, Simon and Varde, Aparna and Weikum, Gerhard. 2022{\natexlab{b}}.
\newblock \href {https://arxiv.org/abs/2210.07763} {\emph{Extracting Cultural Commonsense Knowledge at Scale}}.
\newblock PID \href{https://candle.mpi-inf.mpg.de/}{https://candle.mpi-inf.mpg.de/}.

\bibitem[{Palta and Rudinger(2023)}]{palta2023fork}
Palta, Shramay and Rudinger, Rachel. 2023.
\newblock \href {https://doi.org/10.18653/v1/2023.findings-acl.631} {\emph{{FORK}: A Bite-Sized Test Set for Probing Culinary Cultural Biases in Commonsense Reasoning Models}}.
\newblock Association for Computational Linguistics.
\newblock PID \href{https://github.com/shramay-palta/FORK\_ACL2023}{https://github.com/shramay-palta/FORK\_ACL2023}.

\bibitem[{Romero et~al.(2019)Romero, Razniewski, Pal, Z.~Pan, Sakhadeo, and Weikum}]{romero2019quasimodo}
Romero, Julien and Razniewski, Simon and Pal, Koninika and Z. Pan, Jeff and Sakhadeo, Archit and Weikum, Gerhard. 2019.
\newblock \emph{Commonsense properties from query logs and question answering forums}.
\newblock PID \href{https://quasimodo.mpi-inf.mpg.de/}{https://quasimodo.mpi-inf.mpg.de/}.

\bibitem[{Speer et~al.(2017)Speer, Chin, and Havasi}]{speer2017conceptnet}
Speer, Robyn and Chin, Joshua and Havasi, Catherine. 2017.
\newblock \href {https://doi.org/10.1609/aaai.v31i1.11164} {\emph{Conceptnet 5.5: An open multilingual graph of general knowledge}}.
\newblock PID \href{https://conceptnet.io/}{https://conceptnet.io/}.

\end{thebibliography}


\begin{thebibliography}{57}
\expandafter\ifx\csname natexlab\endcsname\relax\def\natexlab#1{#1}\fi

\bibitem[{Arnaout et~al.(2022)Arnaout, Razniewski, Weikum, and Pan}]{arnaout2022uncommonsense}
Hiba Arnaout, Simon Razniewski, Gerhard Weikum, and Jeff~Z. Pan. 2022.
\newblock \href {https://doi.org/10.1145/3511808.3557484} {Uncommonsense: Informative negative knowledge about everyday concepts}.
\newblock In \emph{Proceedings of the 31st ACM International Conference on Information \& Knowledge Management}, CIKM '22, page 37–46, New York, NY, USA. Association for Computing Machinery.

\bibitem[{Bisk et~al.(2020)Bisk, Zellers, Bras, Gao, and Choi}]{bisk2020piqa}
Yonatan Bisk, Rowan Zellers, Ronan~Le Bras, Jianfeng Gao, and Yejin Choi. 2020.
\newblock \href {https://arxiv.org/abs/1911.11641} {{PIQA}: Reasoning about physical commonsense in natural language}.
\newblock In \emph{Proceedings of the {AAAI} conference on artificial intelligence}, volume~34, pages 7432--7439.

\bibitem[{Bosselut et~al.(2019)Bosselut, Rashkin, Sap, Malaviya, Celikyilmaz, and Choi}]{bosselut2019comet}
Antoine Bosselut, Hannah Rashkin, Maarten Sap, Chaitanya Malaviya, Asli Celikyilmaz, and Yejin Choi. 2019.
\newblock \href {https://doi.org/10.18653/v1/P19-1470} {{COMET}: Commonsense transformers for automatic knowledge graph construction}.
\newblock In \emph{Proceedings of the 57th Annual Meeting of the Association for Computational Linguistics}, pages 4762--4779, Florence, Italy. Association for Computational Linguistics.

\bibitem[{Cachola et~al.(2020)Cachola, Lo, Cohan, and Weld}]{cachola-etal-2020-tldr}
Isabel Cachola, Kyle Lo, Arman Cohan, and Daniel Weld. 2020.
\newblock \href {https://doi.org/10.18653/v1/2020.findings-emnlp.428} {{TLDR}: Extreme summarization of scientific documents}.
\newblock In \emph{Findings of the Association for Computational Linguistics: EMNLP 2020}, pages 4766--4777, Online. Association for Computational Linguistics.

\bibitem[{Cegin et~al.(2023)Cegin, Simko, and Brusilovsky}]{cegin2023chatgpt_diversity}
Jan Cegin, Jakub Simko, and Peter Brusilovsky. 2023.
\newblock Chatgpt to replace crowdsourcing of paraphrases for intent classification: Higher diversity and comparable model robustness.
\newblock \emph{arXiv preprint arXiv:2305.12947}.

\bibitem[{Chandrashekar et~al.(2006)Chandrashekar, Hoon, Ryba, and Zuker}]{chandrashekar2006receptors}
Jayaram Chandrashekar, Mark~A. Hoon, Nicholas J.~P. Ryba, and Charles~S. Zuker. 2006.
\newblock \href {https://doi.org/10.1038/nature05401} {The receptors and cells for mammalian taste}.
\newblock \emph{Nature}, 444(7117):288--294.

\bibitem[{Choi et~al.(2023)Choi, Gim, Badreddine, Kim, Park, and Kang}]{choi2023kitchenscale}
Donghee Choi, Mogan Gim, Samy Badreddine, Hajung Kim, Donghyeon Park, and Jaewoo Kang. 2023.
\newblock \href {https://doi.org/https://doi.org/10.1016/j.eswa.2023.120041} {{KitchenScale}: Learning to predict ingredient quantities from recipe contexts}.
\newblock \emph{Expert Systems with Applications}, 224:120041.

\bibitem[{Chung et~al.(2022)Chung, Hou, Longpre, Zoph, Tay, Fedus, Li, Wang, Dehghani, Brahma, Webson, Gu, Dai, Suzgun, Chen, Chowdhery, Castro-Ros, Pellat, Robinson, Valter, Narang, Mishra, Yu, Zhao, Huang, Dai, Yu, Petrov, Chi, Dean, Devlin, Roberts, Zhou, Le, and Wei}]{chung2022flant5}
Hyung~Won Chung, Le~Hou, Shayne Longpre, Barret Zoph, Yi~Tay, William Fedus, Yunxuan Li, Xuezhi Wang, Mostafa Dehghani, Siddhartha Brahma, Albert Webson, Shixiang~Shane Gu, Zhuyun Dai, Mirac Suzgun, Xinyun Chen, Aakanksha Chowdhery, Alex Castro-Ros, Marie Pellat, Kevin Robinson, Dasha Valter, Sharan Narang, Gaurav Mishra, Adams Yu, Vincent Zhao, Yanping Huang, Andrew Dai, Hongkun Yu, Slav Petrov, Ed~H. Chi, Jeff Dean, Jacob Devlin, Adam Roberts, Denny Zhou, Quoc~V. Le, and Jason Wei. 2022.
\newblock \href {http://arxiv.org/abs/2210.11416v5} {Scaling instruction-finetuned language models}.
\newblock \emph{Computing Research Repository}, arXiv:2210.11416.
\newblock Version 5.

\bibitem[{Davis and Marcus(2015)}]{davis2015commonsense}
Ernest Davis and Gary Marcus. 2015.
\newblock \href {https://doi.org/10.1145/2701413} {Commonsense reasoning and commonsense knowledge in artificial intelligence}.
\newblock \emph{Communications of the ACM}, 58(9):92--103.

\bibitem[{Devlin et~al.(2019)Devlin, Chang, Lee, and Toutanova}]{devlin-etal-2019-bert}
Jacob Devlin, Ming-Wei Chang, Kenton Lee, and Kristina Toutanova. 2019.
\newblock \href {https://doi.org/10.18653/v1/N19-1423} {{BERT}: Pre-training of deep bidirectional transformers for language understanding}.
\newblock In \emph{Proceedings of the 2019 Conference of the North {A}merican Chapter of the Association for Computational Linguistics: Human Language Technologies, Volume 1 (Long and Short Papers)}, pages 4171--4186, Minneapolis, Minnesota. Association for Computational Linguistics.

\bibitem[{Dornenburg and Page(2008)}]{dornenburg2008flavorbible}
Andrew Dornenburg and Karen Page. 2008.
\newblock \emph{The Flavor Bible: The Essential Guide to Culinary Creativity, Based on the Wisdom of America's Most Imaginative Chefs}.
\newblock Little, Brown.

\bibitem[{Ekincek and Günay(2023)}]{ekincek2023recipe_culinary_creativity}
Sema Ekincek and Semra Günay. 2023.
\newblock \href {https://doi.org/https://doi.org/10.1016/j.ijgfs.2022.100633} {A recipe for culinary creativity: Defining characteristics of creative chefs and their process}.
\newblock \emph{International Journal of Gastronomy and Food Science}, 31:100633.

\bibitem[{Fukagawa et~al.(2022)Fukagawa, McKillop, Pehrsson, Moshfegh, Harnly, and Finley}]{fukagawa2022usda}
Naomi~K. Fukagawa, Kyle McKillop, Pamela~R. Pehrsson, Alanna Moshfegh, James Harnly, and John Finley. 2022.
\newblock \href {https://doi.org/10.1093/ajcn/nqab397} {{USDA's FoodData Central}: what is it and why is it needed today?}
\newblock \emph{The American journal of clinical nutrition}, 115(3):619--624.

\bibitem[{Garg et~al.(2017)Garg, Sethupathy, Tuwani, NK, Dokania, Iyer, Gupta, Agrawal, Singh, Shukla, Kathuria, Badhwar, Kanji, Jain, Kaur, Nagpal, and Bagler}]{garg2017flavordb}
Neelansh Garg, Apuroop Sethupathy, Rudraksh Tuwani, Rakhi NK, Shubham Dokania, Arvind Iyer, Ayushi Gupta, Shubhra Agrawal, Navjot Singh, Shubham Shukla, Kriti Kathuria, Rahul Badhwar, Rakesh Kanji, Anupam Jain, Avneet Kaur, Rashmi Nagpal, and Ganesh Bagler. 2017.
\newblock \href {https://doi.org/10.1093/nar/gkx957} {{FlavorDB}: a database of flavor molecules}.
\newblock \emph{Nucleic Acids Research}, 46(D1):D1210--D1216.

\bibitem[{Gim et~al.(2022)Gim, Choi, Maruyama, Choi, Kim, Park, and Kang}]{gim2022recipemind}
Mogan Gim, Donghee Choi, Kana Maruyama, Jihun Choi, Hajung Kim, Donghyeon Park, and Jaewoo Kang. 2022.
\newblock \href {https://doi.org/10.1145/3511808.3557092} {{RecipeMind}: Guiding ingredient choices from food pairing to recipe completion using cascaded set transformer}.
\newblock In \emph{Proceedings of the 31st ACM International Conference on Information \& Knowledge Management}, CIKM '22, page 3092–3102, New York, NY, USA. Association for Computing Machinery.

\bibitem[{Gim et~al.(2021)Gim, Park, Spranger, Maruyama, and Kang}]{gim2021recipebowl}
Mogan Gim, Donghyeon Park, Michael Spranger, Kana Maruyama, and Jaewoo Kang. 2021.
\newblock \href {https://doi.org/10.1109/ACCESS.2021.3120265} {{RecipeBowl}: A cooking recommender for ingredients and recipes using set transformer}.
\newblock \emph{IEEE Access}, 9:143623--143633.

\bibitem[{Grace et~al.(2022)Grace, Finch, Gulbransen-Diaz, and Henderson}]{grace2022qchef}
Kazjon Grace, Elanor Finch, Natalia Gulbransen-Diaz, and Hamish Henderson. 2022.
\newblock \href {https://doi.org/10.1145/3491102.3501862} {{Q-Chef}: The impact of surprise-eliciting systems on food-related decision-making}.
\newblock In \emph{Proceedings of the 2022 CHI Conference on Human Factors in Computing Systems}, CHI '22, New York, NY, USA. Association for Computing Machinery.

\bibitem[{Haussmann et~al.(2019)Haussmann, Seneviratne, Chen, Ne'eman, Codella, Chen, McGuinness, and Zaki}]{haussmann2019foodkg_recommend}
Steven Haussmann, Oshani Seneviratne, Yu~Chen, Yarden Ne'eman, James Codella, Ching-Hua Chen, Deborah~L. McGuinness, and Mohammed~J. Zaki. 2019.
\newblock \href {https://doi.org/10.1007/978-3-030-30796-7_10} {{FoodKG}: a semantics-driven knowledge graph for food recommendation}.
\newblock In \emph{The Semantic Web -- ISWC 2019}, pages 146--162, Cham. Springer International Publishing.

\bibitem[{Hayati et~al.(2023)Hayati, Lee, Rajagopal, and Kang}]{hayati2023diversity}
Shirley~Anugrah Hayati, Minhwa Lee, Dheeraj Rajagopal, and Dongyeop Kang. 2023.
\newblock How far can we extract diverse perspectives from large language models? criteria-based diversity prompting!
\newblock \emph{arXiv preprint arXiv:2311.09799}.

\bibitem[{Honnibal et~al.(2020)Honnibal, Montani, Van~Landeghem, and Boyd}]{spacy2}
Matthew Honnibal, Ines Montani, Sofie Van~Landeghem, and Adriane Boyd. 2020.
\newblock \href {https://doi.org/10.5281/zenodo.1212303} {{spaCy}: Industrial-strength natural language processing in {P}ython}.

\bibitem[{Huang et~al.(2022)Huang, Zhu, Chang, Xiong, and Hwu}]{huang2022deer}
Jie Huang, Kerui Zhu, Kevin Chen-Chuan Chang, Jinjun Xiong, and Wen-mei Hwu. 2022.
\newblock \href {https://aclanthology.org/2022.emnlp-main.448} {{DEER}: Descriptive knowledge graph for explaining entity relationships}.
\newblock In \emph{Proceedings of the 2022 Conference on Empirical Methods in Natural Language Processing}, pages 6686--6698, Abu Dhabi, United Arab Emirates. Association for Computational Linguistics.

\bibitem[{Huang et~al.(2019)Huang, Yu, Chi, Qi, and Xu}]{huang2019healthcare_kg}
Lan Huang, Congcong Yu, Yang Chi, Xiaohui Qi, and Hao Xu. 2019.
\newblock \href {https://doi.org/10.1145/3316615.3316678} {Towards smart healthcare management based on knowledge graph technology}.
\newblock In \emph{Proceedings of the 2019 8th International Conference on Software and Computer Applications}, ICSCA '19, page 330–337, New York, NY, USA. Association for Computing Machinery.

\bibitem[{Hwang et~al.(2021)Hwang, Bhagavatula, Le~Bras, Da, Sakaguchi, Bosselut, and Choi}]{hwang2021comet}
Jena~D. Hwang, Chandra Bhagavatula, Ronan Le~Bras, Jeff Da, Keisuke Sakaguchi, Antoine Bosselut, and Yejin Choi. 2021.
\newblock \href {https://arxiv.org/abs/2010.05953} {{COMET-ATOMIC} 2020: On symbolic and neural commonsense knowledge graphs}.
\newblock In \emph{Proceedings of the {AAAI} Conference on Artificial Intelligence}, volume~35, pages 6384--6392.

\bibitem[{Jiang et~al.(2023)Jiang, Sablayrolles, Mensch, Bamford, Chaplot, Casas, Bressand, Lengyel, Lample, Saulnier et~al.}]{jiang2023mistral}
Albert~Q Jiang, Alexandre Sablayrolles, Arthur Mensch, Chris Bamford, Devendra~Singh Chaplot, Diego de~las Casas, Florian Bressand, Gianna Lengyel, Guillaume Lample, Lucile Saulnier, et~al. 2023.
\newblock Mistral 7b.
\newblock \emph{arXiv preprint arXiv:2310.06825}.

\bibitem[{Johnson and Khoshgoftaar(2019)}]{johnson2019survey_classimbalance}
Justin~M. Johnson and Taghi~M. Khoshgoftaar. 2019.
\newblock \href {https://doi.org/10.1186/s40537-019-0192-5} {Survey on deep learning with class imbalance}.
\newblock \emph{Journal of Big Data}, 6(27).

\bibitem[{Kim et~al.(2023)Kim, Park, Kwon, Jo, Thorne, and Choi}]{kim2023factkg}
Jiho Kim, Sungjin Park, Yeonsu Kwon, Yohan Jo, James Thorne, and Edward Choi. 2023.
\newblock \href {https://doi.org/10.18653/v1/2023.acl-long.895} {{F}act{KG}: Fact verification via reasoning on knowledge graphs}.
\newblock In \emph{Proceedings of the 61st Annual Meeting of the Association for Computational Linguistics (Volume 1: Long Papers)}, pages 16190--16206, Toronto, Canada. Association for Computational Linguistics.

\bibitem[{Lei et~al.(2021)Lei, Haq, Zeb, Suzauddola, and Zhang}]{lei2021rcpkg}
Zhenfeng Lei, Anwar~Ul Haq, Adnan Zeb, Md~Suzauddola, and Defu Zhang. 2021.
\newblock \href {https://doi.org/10.1016/j.eswa.2021.115708} {Is the suggested food your desired?: Multi-modal recipe recommendation with demand-based knowledge graph}.
\newblock \emph{Expert Systems with Applications}, 186:115708.

\bibitem[{Lewis et~al.(2020{\natexlab{a}})Lewis, Liu, Goyal, Ghazvininejad, Mohamed, Levy, Stoyanov, and Zettlemoyer}]{lewis2019bart}
Mike Lewis, Yinhan Liu, Naman Goyal, Marjan Ghazvininejad, Abdelrahman Mohamed, Omer Levy, Veselin Stoyanov, and Luke Zettlemoyer. 2020{\natexlab{a}}.
\newblock \href {https://doi.org/10.18653/v1/2020.acl-main.703} {{BART}: Denoising sequence-to-sequence pre-training for natural language generation, translation, and comprehension}.
\newblock In \emph{Proceedings of the 58th Annual Meeting of the Association for Computational Linguistics}, pages 7871--7880, Online. Association for Computational Linguistics.

\bibitem[{Lewis et~al.(2020{\natexlab{b}})Lewis, Perez, Piktus, Petroni, Karpukhin, Goyal, K\"{u}ttler, Lewis, Yih, Rockt\"{a}schel, Riedel, and Kiela}]{lewis2020rag}
Patrick Lewis, Ethan Perez, Aleksandra Piktus, Fabio Petroni, Vladimir Karpukhin, Naman Goyal, Heinrich K\"{u}ttler, Mike Lewis, Wen-tau Yih, Tim Rockt\"{a}schel, Sebastian Riedel, and Douwe Kiela. 2020{\natexlab{b}}.
\newblock \href {https://proceedings.neurips.cc/paper_files/paper/2020/file/6b493230205f780e1bc26945df7481e5-Paper.pdf} {Retrieval-augmented generation for knowledge-intensive {NLP} tasks}.
\newblock In \emph{Advances in Neural Information Processing Systems}, volume~33, pages 9459--9474. Curran Associates, Inc.

\bibitem[{Lo et~al.(2020)Lo, Wang, Neumann, Kinney, and Weld}]{lo2020s2orc}
Kyle Lo, Lucy~Lu Wang, Mark Neumann, Rodney Kinney, and Daniel Weld. 2020.
\newblock \href {https://doi.org/10.18653/v1/2020.acl-main.447} {{S}2{ORC}: The semantic scholar open research corpus}.
\newblock In \emph{Proceedings of the 58th Annual Meeting of the Association for Computational Linguistics}, pages 4969--4983, Online. Association for Computational Linguistics.

\bibitem[{L{\'o}pez-Alt(2015)}]{lopez2015foodlab}
J.~Kenji L{\'o}pez-Alt. 2015.
\newblock \emph{The food lab: better home cooking through science}.
\newblock WW Norton \& Company.

\bibitem[{Marcus(2013)}]{marcus2013culinary_nutrition}
Jacqueline~B. Marcus. 2013.
\newblock \emph{Culinary nutrition: the science and practice of healthy cooking}.
\newblock Academic Press.

\bibitem[{Marin et~al.(2019)Marin, Biswas, Ofli, Hynes, Salvador, Aytar, Weber, and Torralba}]{marin2019recipe1m+}
Javier Marin, Aritro Biswas, Ferda Ofli, Nicholas Hynes, Amaia Salvador, Yusuf Aytar, Ingmar Weber, and Antonio Torralba. 2019.
\newblock \href {https://arxiv.org/abs/1810.06553} {{Recipe1M+}: A dataset for learning cross-modal embeddings for cooking recipes and food images}.
\newblock \emph{{IEEE} Transactions on Pattern Analysis and Machine Intelligence}, 43(1):187--203.

\bibitem[{Min et~al.(2022)Min, Liu, Xu, and Jiang}]{min2022foodkg_survey}
Weiqing Min, Chunlin Liu, Leyi Xu, and Shuqiang Jiang. 2022.
\newblock \href {https://doi.org/https://doi.org/10.1016/j.patter.2022.100484} {Applications of knowledge graphs for food science and industry}.
\newblock \emph{Patterns}, 3(5):100484.

\bibitem[{Nguyen et~al.(2021)Nguyen, Razniewski, and Weikum}]{nguyen2021ascent}
Tuan-Phong Nguyen, Simon Razniewski, and Gerhard Weikum. 2021.
\newblock \href {https://doi.org/10.1145/3442381.3449827} {Advanced semantics for commonsense knowledge extraction}.
\newblock In \emph{Proceedings of the Web Conference 2021}, WWW '21, page 2636–2647, New York, NY, USA. Association for Computing Machinery.

\bibitem[{Nian et~al.(2021)Nian, Du, Bu, Li, Hu, Zhang, and Tao}]{nian2021disease_discovery}
Yi~Nian, Jingcheng Du, Larry Bu, Fang Li, Xinyue Hu, Yuji Zhang, and Cui Tao. 2021.
\newblock \href {http://arxiv.org/abs/2109.06123v2} {Knowledge graph-based neurodegenerative diseases and diet relationship discovery}.
\newblock \emph{Computing Research Repository}, arXiv:2109.06123.
\newblock Version 2.

\bibitem[{Nori et~al.(2023)Nori, Lee, Zhang, Carignan, Edgar, Fusi, King, Larson, Li, Liu et~al.}]{nori2023medprompt}
Harsha Nori, Yin~Tat Lee, Sheng Zhang, Dean Carignan, Richard Edgar, Nicolo Fusi, Nicholas King, Jonathan Larson, Yuanzhi Li, Weishung Liu, et~al. 2023.
\newblock Can generalist foundation models outcompete special-purpose tuning? case study in medicine.
\newblock \emph{arXiv preprint arXiv:2311.16452}.

\bibitem[{Ouyang et~al.(2022)Ouyang, Wu, Jiang, Almeida, Wainwright, Mishkin, Zhang, Agarwal, Slama, Ray, Schulman, Hilton, Kelton, Miller, Simens, Askell, Welinder, Christiano, Leike, and Lowe}]{ouyang2022training}
Long Ouyang, Jeffrey Wu, Xu~Jiang, Diogo Almeida, Carroll Wainwright, Pamela Mishkin, Chong Zhang, Sandhini Agarwal, Katarina Slama, Alex Ray, John Schulman, Jacob Hilton, Fraser Kelton, Luke Miller, Maddie Simens, Amanda Askell, Peter Welinder, Paul~F Christiano, Jan Leike, and Ryan Lowe. 2022.
\newblock \href {https://proceedings.neurips.cc/paper_files/paper/2022/file/b1efde53be364a73914f58805a001731-Paper-Conference.pdf} {Training language models to follow instructions with human feedback}.
\newblock In \emph{Advances in Neural Information Processing Systems}, volume~35, pages 27730--27744. Curran Associates, Inc.

\bibitem[{Park et~al.(2021)Park, Kim, Kim, Spranger, and Kang}]{park2021flavorgraph}
Donghyeon Park, Keonwoo Kim, Seoyoon Kim, Michael Spranger, and Jaewoo Kang. 2021.
\newblock \href {https://doi.org/10.1038/s41598-020-79422-8} {{FlavorGraph}: a large-scale food-chemical graph for generating food representations and recommending food pairings}.
\newblock \emph{Scientific Reports}, 11(931).

\bibitem[{Park et~al.(2019)Park, Kim, Park, Shin, and Kang}]{park2019kitchenette}
Donghyeon Park, Keonwoo Kim, Yonggyu Park, Jungwoon Shin, and Jaewoo Kang. 2019.
\newblock \href {https://doi.org/10.24963/ijcai.2019/822} {{KitcheNette}: Predicting and ranking food ingredient pairings using siamese neural network}.
\newblock In \emph{Proceedings of the Twenty-Eighth International Joint Conference on Artificial Intelligence, {IJCAI-19}}, pages 5930--5936. International Joint Conferences on Artificial Intelligence Organization.

\bibitem[{Pizzuti and Mirabelli(2013)}]{pizzuti2013ftto}
Teresa Pizzuti and Giovanni Mirabelli. 2013.
\newblock \href {https://doi.org/10.1109/IDAACS.2013.6662689} {{FTTO}: an example of food ontology for traceability purpose}.
\newblock In \emph{2013 IEEE 7th International Conference on Intelligent Data Acquisition and Advanced Computing Systems (IDAACS)}, volume~1, pages 281--286. IEEE.

\bibitem[{Raffel et~al.(2020)Raffel, Shazeer, Roberts, Lee, Narang, Matena, Zhou, Li, and Liu}]{raffel2020t5}
Colin Raffel, Noam Shazeer, Adam Roberts, Katherine Lee, Sharan Narang, Michael Matena, Yanqi Zhou, Wei Li, and Peter~J. Liu. 2020.
\newblock \href {http://jmlr.org/papers/v21/20-074.html} {Exploring the limits of transfer learning with a unified text-to-text transformer}.
\newblock \emph{Journal of Machine Learning Research}, 21(140):1--67.

\bibitem[{Rehman et~al.(2017)Rehman, Khalid, ul~Haq, ur~Rehman~Khan, Bilal, and Madani}]{rehman2017diet}
Faisal Rehman, Osman Khalid, Nuhman ul~Haq, Atta ur~Rehman~Khan, Kashif Bilal, and Sajjad~A. Madani. 2017.
\newblock \href {https://doi.org/10.3837/tiis.2017.06.006} {{Diet-Right}: A smart food recommendation system}.
\newblock \emph{KSII Transactions on Internet and Information Systems}, 11(6):2910--2925.

\bibitem[{Robertson et~al.(2009)Robertson, Zaragoza et~al.}]{robertson2009bm25}
Stephen Robertson, Hugo Zaragoza, et~al. 2009.
\newblock \href {https://doi.org/10.1561/1500000019} {The probabilistic relevance framework: {BM25} and beyond}.
\newblock \emph{Foundations and Trends{\textregistered} in Information Retrieval}, 3(4):333--389.

\bibitem[{Sakib et~al.(2023)Sakib, Shahariar, Kabir, Hasan, and Mahmud}]{sakib2023_3a2m}
Nazmus Sakib, G.~M. Shahariar, Md.~Mohsinul Kabir, Md.~Kamrul Hasan, and Hasan Mahmud. 2023.
\newblock \href {https://doi.org/10.1007/978-3-031-34622-4_15} {Assorted, archetypal and annotated two million ({3A2M}) cooking recipes dataset based on active learning}.
\newblock In \emph{Machine Intelligence and Emerging Technologies}, pages 188--203, Cham. Springer Nature Switzerland.

\bibitem[{Sanh et~al.(2022)Sanh, Webson, Raffel, Bach, Sutawika, Alyafeai, Chaffin, Stiegler, Raja, Dey, Bari, Xu, Thakker, Sharma, Szczechla, Kim, Chhablani, Nayak, Datta, Chang, Jiang, Wang, Manica, Shen, Yong, Pandey, Bawden, Wang, Neeraj, Rozen, Sharma, Santilli, Fevry, Fries, Teehan, Scao, Biderman, Gao, Wolf, and Rush}]{sanh2022multitask}
Victor Sanh, Albert Webson, Colin Raffel, Stephen Bach, Lintang Sutawika, Zaid Alyafeai, Antoine Chaffin, Arnaud Stiegler, Arun Raja, Manan Dey, M~Saiful Bari, Canwen Xu, Urmish Thakker, Shanya~Sharma Sharma, Eliza Szczechla, Taewoon Kim, Gunjan Chhablani, Nihal Nayak, Debajyoti Datta, Jonathan Chang, Mike Tian-Jian Jiang, Han Wang, Matteo Manica, Sheng Shen, Zheng~Xin Yong, Harshit Pandey, Rachel Bawden, Thomas Wang, Trishala Neeraj, Jos Rozen, Abheesht Sharma, Andrea Santilli, Thibault Fevry, Jason~Alan Fries, Ryan Teehan, Teven~Le Scao, Stella Biderman, Leo Gao, Thomas Wolf, and Alexander~M. Rush. 2022.
\newblock \href {https://openreview.net/forum?id=9Vrb9D0WI4} {Multitask prompted training enables zero-shot task generalization}.
\newblock In \emph{International Conference on Learning Representations}.

\bibitem[{Snae and Bruckner(2008)}]{snae2008foods}
Chakkrit Snae and Michael Bruckner. 2008.
\newblock \href {https://doi.org/10.1109/DEST.2008.4635195} {{FOODS}: a food-oriented ontology-driven system}.
\newblock In \emph{2008 2nd ieee international conference on digital ecosystems and technologies}, pages 168--176. IEEE.

\bibitem[{Spencer et~al.(2017)Spencer, Korosi, Lay{\'e}, Shukitt-Hale, and Barrientos}]{spencer2017food}
Sarah~J. Spencer, Aniko Korosi, Sophie Lay{\'e}, Barbara Shukitt-Hale, and Ruth~M. Barrientos. 2017.
\newblock \href {https://doi.org/10.1038/s41538-017-0008-y} {Food for thought: how nutrition impacts cognition and emotion}.
\newblock \emph{npj Science of Food}, 1(1):7.

\bibitem[{Takeshita et~al.(2022)Takeshita, Green, Friedrich, Eckert, and Ponzetto}]{takeshita2022x-scitldr}
Sotaro Takeshita, Tommaso Green, Niklas Friedrich, Kai Eckert, and Simone~Paolo Ponzetto. 2022.
\newblock X-scitldr: cross-lingual extreme summarization of scholarly documents.
\newblock In \emph{Proceedings of the 22nd ACM/IEEE Joint Conference on Digital Libraries}, pages 1--12.

\bibitem[{Team et~al.(2024)Team, Mesnard, Hardin, Dadashi, Bhupatiraju, Pathak, Sifre, Rivi{\`e}re, Kale, Love et~al.}]{team2024gemma}
Gemma Team, Thomas Mesnard, Cassidy Hardin, Robert Dadashi, Surya Bhupatiraju, Shreya Pathak, Laurent Sifre, Morgane Rivi{\`e}re, Mihir~Sanjay Kale, Juliette Love, et~al. 2024.
\newblock Gemma: Open models based on gemini research and technology.
\newblock \emph{arXiv preprint arXiv:2403.08295}.

\bibitem[{Touvron et~al.(2023)Touvron, Martin, Stone, Albert, Almahairi, Babaei, Bashlykov, Batra, Bhargava, Bhosale et~al.}]{touvron2023llama2}
Hugo Touvron, Louis Martin, Kevin Stone, Peter Albert, Amjad Almahairi, Yasmine Babaei, Nikolay Bashlykov, Soumya Batra, Prajjwal Bhargava, Shruti Bhosale, et~al. 2023.
\newblock Llama 2: Open foundation and fine-tuned chat models.
\newblock \emph{arXiv preprint arXiv:2307.09288}.

\bibitem[{Trotman et~al.(2014)Trotman, Puurula, and Burgess}]{trotman2014improvements}
Andrew Trotman, Antti Puurula, and Blake Burgess. 2014.
\newblock \href {https://doi.org/10.1145/2682862.2682863} {Improvements to bm25 and language models examined}.
\newblock In \emph{Proceedings of the 19th Australasian Document Computing Symposium}, ADCS '14, page 58–65, New York, NY, USA. Association for Computing Machinery.

\bibitem[{Wang et~al.(2023)Wang, Liu, Liu, Jiang, Li, and Xiao}]{wang2023sweet_commonsense}
Chao Wang, Juntao Liu, Jingping Liu, Sihang Jiang, Zhixu Li, and Yanghua Xiao. 2023.
\newblock \href {https://doi.org/https://doi.org/10.1016/j.knosys.2023.110988} {Sweet apple, company? or food? adjective-centric commonsense knowledge acquisition with taxonomy-guided induction}.
\newblock \emph{Knowledge-Based Systems}, 280:110988.

\bibitem[{Wei et~al.(2022{\natexlab{a}})Wei, Bosma, Zhao, Guu, Yu, Lester, Du, Dai, and Le}]{wei2022finetuned}
Jason Wei, Maarten Bosma, Vincent Zhao, Kelvin Guu, Adams~Wei Yu, Brian Lester, Nan Du, Andrew~M. Dai, and Quoc~V. Le. 2022{\natexlab{a}}.
\newblock \href {https://openreview.net/forum?id=gEZrGCozdqR} {Finetuned language models are zero-shot learners}.
\newblock In \emph{International Conference on Learning Representations}.

\bibitem[{Wei et~al.(2022{\natexlab{b}})Wei, Wang, Schuurmans, Bosma, Xia, Chi, Le, Zhou et~al.}]{wei2022cot}
Jason Wei, Xuezhi Wang, Dale Schuurmans, Maarten Bosma, Fei Xia, Ed~Chi, Quoc~V Le, Denny Zhou, et~al. 2022{\natexlab{b}}.
\newblock Chain-of-thought prompting elicits reasoning in large language models.
\newblock \emph{Advances in neural information processing systems}, 35:24824--24837.

\bibitem[{Williams et~al.(2018)Williams, Nangia, and Bowman}]{williams2018multinli}
Adina Williams, Nikita Nangia, and Samuel Bowman. 2018.
\newblock \href {http://aclweb.org/anthology/N18-1101} {A broad-coverage challenge corpus for sentence understanding through inference}.
\newblock In \emph{Proceedings of the 2018 Conference of the North American Chapter of the Association for Computational Linguistics: Human Language Technologies, Volume 1 (Long Papers)}, pages 1112--1122. Association for Computational Linguistics.

\bibitem[{Zhang et~al.(2023)Zhang, Gao, Zheng, Fan, Lai, Zhang, Ai, Yang, and Yang}]{zhang2023domainchatbot}
Ruohong Zhang, Luyu Gao, Chen Zheng, Zhen Fan, Guokun Lai, Zheng Zhang, Fangzhou Ai, Yiming Yang, and Hongxia Yang. 2023.
\newblock A self-enhancement approach for domain-specific chatbot training via knowledge mining and digest.
\newblock \emph{arXiv preprint arXiv:2311.10614}.

\end{thebibliography}

\section{Language Resource References}
\label{lr:ref}
\bibliographystylelanguageresource{lrec-coling2024-natbib}
\bibliographylanguageresource{languageresource}

\end{document}